\documentclass{article}

\usepackage{algorithm}
\usepackage{arxiv}
\usepackage{graphicx}
\usepackage[numbers]{natbib}
\usepackage[svgnames]{xcolor}
\usepackage{algorithm}
\usepackage{graphicx}
\usepackage{graphicx}
\usepackage{algcompatible}
\usepackage{amsmath}
\usepackage{amssymb}
\usepackage{caption}
\usepackage{balance}
\usepackage{listings}
\usepackage{color}
\usepackage{float}
\usepackage[utf8]{inputenc} 
\usepackage[T1]{fontenc}    
\usepackage{hyperref}       
\usepackage{url}            
\usepackage{booktabs}       
\usepackage{amsfonts}       
\usepackage{nicefrac}       
\usepackage{microtype}      
\usepackage{lipsum}		
\usepackage{graphicx}
\usepackage{natbib}
\usepackage{doi}

\title{Unsupervised Technique To Conversational Machine Reading}

\author{ \href{https://orcid.org/0000-0000-0000-0000}{\includegraphics[scale=0.06]{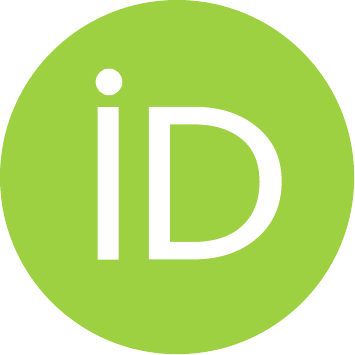}\hspace{1mm}Peter Ochieng}\thanks{} \\
	Department of Computing and Informatics\\
	Taita Taveta University\\
	P.O. Box 635 – 80300
Voi,Kenya. \\
	\texttt{peter.ochieng@ttu.ac.ke} \\
	\And
	\href{https://orcid.org/0000-0000-0000-0000}{\includegraphics[scale=0.06]{orcid.pdf}\hspace{1mm}Dennis Kaburu} \\
	Department of Computing\\
	Jomo Kenyatta University of Agriculture and Technology\\
	P.O. Box 62000 CITY SQUARE,Nairobi
,Kenya.
	 \\
	\texttt{dennis.kaburu@jkuat.ac.ke } \\
}

\renewcommand{\shorttitle}{\textit{}}

\hypersetup{
pdftitle={A template for the arxiv style},
pdfsubject={q-bio.NC, q-bio.QM},
pdfauthor={Peter Ochieng, Dennis Kaburu},
pdfkeywords={First keyword, Second keyword, More},
}

\begin{document}
\maketitle

\begin{abstract}
	Conversational machine reading (CMR) tools have seen a rapid progress in the recent past. The current existing tools rely  on the supervised learning technique which require labeled dataset for their training. The supervised technique necessitates  that for every new rule text, a manually  labeled dataset must be created. This is tedious and error prone. This paper introduces and demonstrates how unsupervised learning technique can be applied in the development of CMR. Specifically, we demonstrate how unsupervised learning can be used in rule extraction and entailment modules of CMR. Compared to the current best CMR tool, our developed framework reports 3.3\% improvement in micro averaged accuracy and 1.4 \% improvement in 
macro averaged accuracy.
\end{abstract}

\keywords{Chatbot \and Conversational\and NLP\and Natural Language Processing\and Dialog\and Unsupervised Learning}

\section{Introduction}
Conversational machine reading (CMR) tools allow users to give a description of their scenario and pose a question to them \cite{review2020}  \cite{E2020}. The CMR tool then processes the rule text in relation to the user scenario and question and either picks an  appropriate answer from the set \sloppy of possible answers $\mathcal{A}=\textrm{\{Yes, No, Irrelevant\}}$ or chooses to seek futher clarification before giving an answer from the set $\mathcal{A}$ \cite{discourse2020}. A number of systems  \cite{E2020} \cite{discourse2020}   \cite{dialog2020}  \cite{seg2018} have been developed with a goal to improve the precision of the answers given to the user. However, all the existing tools apply supervised learning technique which require  manually labeled dataset. For every new rule text, the supervised techniques will require that a labeled dataset be created.  The  task of manually labeling dataset is tedious and error prone  \cite{Alonso2015}. Moreover, it  may not generate enough dataset for proper training of the developed model. Due to limited training dataset, the model runs the risk of performing  over fitting. Unsupervised learning has shown remarkable success in other fields such as machine translation \cite{Lample2018}. They have attained this success using no labeled training data. Motivated by this,  we also introduce  unsupervised learning technique as part of the CMR tool.
\par Further, the existing systems take the view that they are interacting with a knowledgeable user who has some basic idea of the  subject he or she wants to inquire about. Take an example of a farmer keeping chicken who has observed a number of symptoms in his or her chicken. He or she will describe a scenario as  shown on the left side of  Fig 2.
\begin{figure}[ht]
	\centering
\includegraphics[scale=0.67,angle=0]{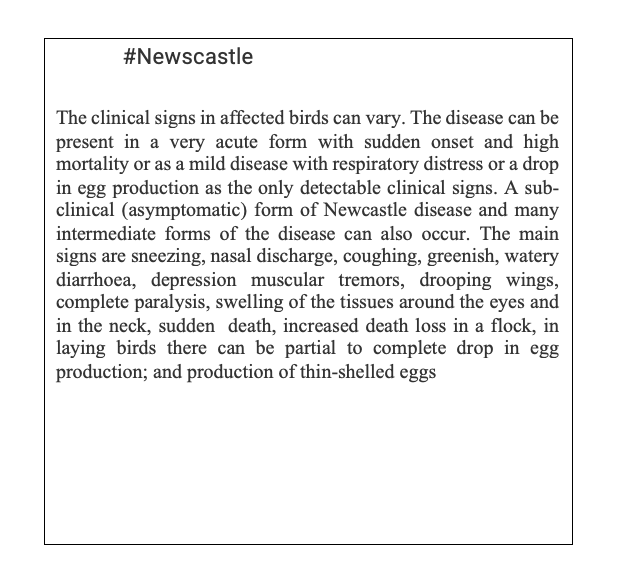}
		\caption{Sample rule text  }
	\label{fig:example2}
\end{figure}
\begin{figure}[ht]
	\centering
		\includegraphics[scale=0.5,angle=0]{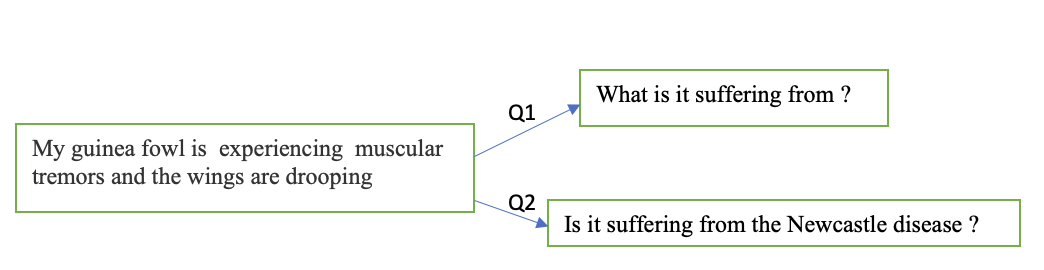}
		\caption{User Scenario and possible questions }
	\label{fig:example2}
\end{figure}
It takes a user who has some prior knowledge about chicken diseases to pose question 2 (Q2) to the CMR system which processes the rule text of chicken diseases ( a snippet is shown in figure 1). However, for a novice in chicken farming question 1 (Q1) is the most  likely. Therefore, there is a need to develop a CMR system that is able to handle both specific questions such as Q2 and general question such Q1. This work proposes a new 
CMR system that makes the following key contribution:
\begin{enumerate}
\item We demonstrate how unsupervised learning technique can be exploited for both rule  extraction and  entailment modules in the CMR system. 
\item We develop A CMR tool that can handle both specific and general questions
\end{enumerate}
Experiments  conducted on the ShARC dataset      \cite{Saeidi2020} demonstrate that the proposed approach provides a state of the art results.

\section{Related Work}
CMR systems are always built as an aggregation of different modules with each module exploiting a given technology \cite{review2020} \cite{E2020}.  In this section we focus the discussion on  the techniques the existing CMR tools exploit to implement the two key modules of CMR i.e the rule extraction and rule entailment modules.
Rule extraction module implements techniques to extract rules $\{r_1,r_2,\cdots, r_n\}$ of a given subject from the rule text.  Given the rule set $\{r_1,r_2,\cdots r_n\}$ extracted from a rule text, the rule entailment module seeks to check whether a given rule $r_i$ is entailed in the conversation history.
\par To extract rules from the rule text, work in \cite{dialog2020} first extracts elementary discourse units (EDUs). The extracted EDUs are then exploited to construct an explicit discourse graph. To establish a link between the rule text and user scenario,  a user scenario representation is fed into the explicit graph as  global vertex. Further,  a second implicit discourse graph is designed for extracting the latent salient interactions among rule texts.  The two  graphs are then exploited  for making decisions.
In order to process the rule  text and to extract  rules from it, \cite{discourse2020} proposes to segment the rule text into  elementary discourse units (EDUs) using a pretrained discourse segmentation model proposed by \cite{seg2018}. Each EDU is then treated as a condition of the rule text. Similar to \cite{discourse2020}, the work in \cite{up2021} also uses discourse segmentation to extract rules in the rule text. Work in \cite{E2020} first uses  Bidirectional Encoder Representations from Transformers (BERT) \cite{BERT2020} to encode the text in  rule text, user scenario, user inquiry and system inquiry. It then uses attention based heuristics to extract rules that exist in the rule text. 
\par Given a set of rules and  a sequence of user-provided information, \cite{discourse2020}  utilizes  the transformer encoder \cite{trans2020} to predict the entailment states for all the rules. The transformer  outputs whether a rule is an entailment, a contradicion or a neutral. In \cite{dialog2020},  once all the EDUs have been extracted,  they train a model   via a cross entropy loss to perform a multi-class classification with regards to entailment of a given EDU. To check for entailment,  \cite{up2021} processed the  ShARC dataset to  generate a training dataset where each EDU is linked with its most similar dialog history. They then manually label the linked pair using the tags "Entailment"  if the answer for the mentioned follow-up question is a Yes ,  Contradiction" if the answer is a No or "Neural"  if  the EDUs is not matched to any follow-up question. A model is the trained using a supervised training technique to recognise the three classes. The work in \cite{E2020} establishes  rule entailment by   computing  a similarity  score that exploits the number of shared  tokens  between the dialog history and the extracted rule. 
\section{Unsupervised based Conversational Machine Reading Tool}
\subsection{Rule Extraction}

\begin{figure}[ht]
	\centering
\includegraphics[scale=0.67,angle=0]{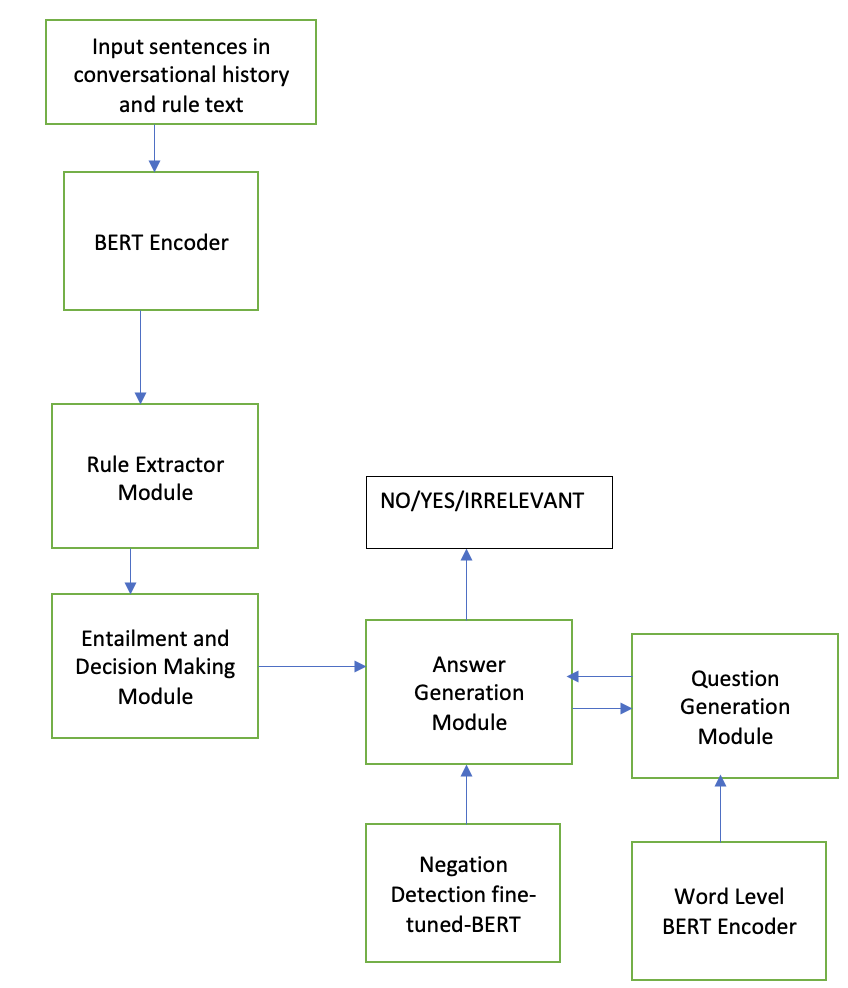}
		\caption{The overall structure of the proposed Unsupervised based Conversational Machine Reading Tool (UCMRT)  }
	\label{fig:example2}
\end{figure}
\par In a  rule text, a given set of  span of  sentences contain a set of  latent rules linked to a given topic or subject. This module   first extracts a span of sentences in the rule text that address a given topic or subject. Then from this set of sentences the module extracts the latent rules that are related to the topic (subject). The module outputs  a number of rule sets where each set contains rules that relate to a given subject or topic. As an initial step,
the  module  encodes  sentences in  a context aware manner. To do this, we exploit BERT  which encodes  the sentences  in  the rule text ($RT$), user scenario ($US$), user question ($UQ$),  for each  $i^{th}$ turn in the conversation, the system's inquiry is concatenated  with the user response to form a single  inquiry response($IR$)  sentence. The sentences are  then structured to  construct BERT  input sequence as  [ [CLS]  Sentence1  [SEP] Sentence2 ].  The input is then fed into BERT which first tokenizes the words in a sentence  using WordPiece tokenizer. The tokenized words are then  embedded  with both their positional embeddings and segmentation embeddings. These embeddings are subsequently encoded via a transformer network. The output of BERT encoder $R^{H}$  is a vector representing an  encoding  of a  sentence $S_i$.

For each sentence $S_i$, we seek to establish a set of   related sentences $S_{relatives_i}$  such that 
$\forall S_j\in S_{relatives_i}, |j-i|\geq1$
where $i$ and $j$ are position of a sentence in the rule text.
To create the set $ S_{relatives_i}$  of  $S_i$, we use dissimilarity score
\begin{equation}
DisScore(S_i,S_{i+1})=-sim(S_i,S_{i+1})
\end{equation}
where $i=1,2,\cdots,L-1$ where $L$ is the number of sentences in the rule set.
Intuitively, $DisScore(S_i,S_{i+1})$ evaluates the  confidence that the next subsequent sentence $S_{i+1}$ addresses  a  different topic (subject)  from the current  sentence $S_i$. Thus  sentences positions  with high dissimilarity values is a signal of subject   change within the rule text, and are considered as candidates for subject  change boundary. To extract the set of  all $ S_{relatives_i}$ within a rule text, we apply a peak detection algorithm over the dissimilarity values,    $DisScore(S_i,S_{i+1})$. The $DisScore(S_i,S_{i+1})$ for which the score exceeds a peak set threshold  $\vartheta$ are predicted as boundaries. 
\subsubsection{ Extracting latent rules within a subject.}
For a given set  $ S_{relatives_i}$ that contains sentences addressing a given subject, the   latent rules $r_1,r_2,\cdots,r_t$ are established. To do this,  an undirected graph $G$  is created with vertex and edge set V(G)  and E(G) respectively. Each $S_j\in S_{relatives_i} $ is a vertex $v_j\in V(G)$. Each edge $e_i\in E(G)$ is weighted  using a weighting function $w$
$w: V(G)\times V(G)\mapsto R^{+}$ using the similarity score
\begin{equation}
SimScore(S_i,S_j)=sim(S_i,S_j)
\end{equation}
From $G$, we extract a weighted  adjacent matrix $W=w_{ij} $ where $i,j=1,2,\cdots ,|S_{relatives_i}|$.
The degree $d_i$ of a vertex $v_i\in V(G)$ is defined as: 
\begin{equation}
 d_i = \sum_{n=1}^{n} w_{ij}
\end{equation}
We then define the degree matrix $D$ as a diagonal matrix with degrees $d_i,\cdots , d_n$ on the diagonal. Using $D$, we then construct a Laplacian matrix $L$. 

\begin{equation}
L= D-W
\end{equation}
L has a number of properties \cite{Mohar1988} \cite{Mohar1991}. The properties of interest to this work is that the unnormalized graph Laplacian matrix $L$,  its eigenvalues and eigenvectors can be used  for spectral clustering of the sentences within  $ S_{relatives_i} $. We hypothesize that the different clusters of $ S_{relatives_i} $ represent different set of rules contained within a subject. We therefore use the spectral clustering shown in Algorithm 1 to extract distinct rules contained in $ S_{relatives_i} $. 

\begin{algorithm}
\caption{Rule extraction based on spectral clustering algorithm.}
\begin{algorithmic}[1]\\
\textbf{Input:} $k$ number of clusters to be constructed and $ S_{relatives_i}$
\State Construct similarity matrix graph $G$ as described in section 3.1.
\State Construct the weighted adjacent matrix w for the graph $G$.
\State Construct the unnormalized Laplacian matrix L as shown in equation 3.
\State Compute the first $k$ eigenvalues and the corresponding eigenvectors $u_1,u_2,\cdots u_n$.
\State let $U\in R^{H\times k} $ be a matrix constituting the vectors $u_1,u_2,\cdots,u_k $ as columns.
\State for $i=1,\cdots,n$ let $y_i\in R^{k}$ be a vector corresponding to the $i^{th}$ row of $U$.
\State Cluster the points $y_i$ $i=1,\cdots, n$ in $R^k$ with the k-means algorithm into clusters $c_1,\cdots,c_k$.
 \State  Clusters $ K_1,K_2\cdots K_k$ with $K_1=\{r_j|y_j\in C_i\}$.
 \State Merge the clusters $ K_1,K_2\cdots K_k $ into a single set $R_i$ containing all  distinct rules $r_i$  from the clusters $ K_1,K_2\cdots K_k$
\textbf{Output}: The set  $R_i$ containing all rules  in  $ S_{relatives_i}$

\end{algorithmic}
\end{algorithm}
The module finally creates a universal set  $\mathcal{U}= \{r_1,r_2,\cdots,r_n$\} containing all the rules extracted in the rule text. Further, a set $\mathcal{Q}= \{R_1,R_2,\cdots,R_n$\} is created  such that $ \mathcal{Q}$ contains all rule sets contained in the rule text extracted by Algorithm 1. 
\section{Rule Entailment}
The entailment module seeks to establish whether the conversational history  fully covers all the rules in set $R_i\in \mathcal{Q}$  or some rules are still left out. This helps the system to make a decision on whether to seek further clarification based on the uncovered rules  or give a definitive answer to the user. Here,   a Generative Adversarial Network (GAN) model \cite{Goodfellow2016} is set up which is able to output a set  of rules 	$\mathcal{P}$ given a certain span of sentences.  The  GAN is an unsupervised  model  that constitutes two key parts i.e the generator $\mathcal{G}$ and the discriminator $\mathcal{D}$, where $\mathcal{G}$  generates samples which are then  judged by the $\mathcal{D}$. The discriminator $\mathcal{D}$ is trained to classify whether samples are  from a real data distribution or not. The objective of the generator is to produce samples that can appear to the discriminator  as  data from real data distribution.
\par In this work, $\mathcal{G}$ takes as  its input a span  of $\mathcal{L}$ sentences representation i.e $\mathcal{S}=\{S_1,S_2,\cdots,S_{\mathcal{L}}\}$. It then maps the span of sentences representation  to a sequence of $\mathcal{K}$ rules  $\{r_1,r_2,\cdots,r_{\mathcal{K}}\}$. The generator essentially predicts the probability distribution  over the universal set $\mathcal{U}$ for each span of sentences  $\mathcal{L}$  and outputs a set of rules $\mathcal{P}$ $\subseteq \mathcal{U}$  with the highest probability. The generator's output has a dimension of $|\mathcal{U}|$ in form of  one hot encoding   with 1 indicating that rule $r_i$ is contained  in the span  of $\mathcal{L}$ sentences while 0 indicates that rule $r_i$ is not contained in the  $\mathcal{L}$ input sentences.
\par The  discriminator $\mathcal{D}$ on the other hand  takes as an input $|\mathcal{U}|$ dimension one hot encoding  of a set of rules $\mathcal{P}^x$ representing  a set  either from  $\mathcal{P}$ i.e from the generator  or $R_i\in \mathcal{Q}$  i.e from  real set of rules.
  The discriminator $\mathcal{D}$ outputs a probability indicating the likelihood  that the sample  from the  real rule set. This work uses the objective as original
as proposed by \cite{Goodfellow2016} \cite{Goodfellow2020} 
\begin{equation}
 \min_{\mathcal{G}} \max_{\mathcal{D}}=
\mathbb{E}_{\mathcal{P}^x\sim\mathcal{P}^x}[\log D(\mathcal{P}^x)] -\mathbb{E}_{\mathcal{S}\sim \mathcal{S}}[\log (1 - D(G(\mathcal{S})))-
 \beta \mathcal{G}_p+\theta \mathcal{S}_p
\end{equation}
Here,  $\mathcal{P}^x$ represents a rule set from the set  $\mathcal{Q}$ while  $\mathcal{G(S)}$ is the generated set of rules $\mathcal{P}$ produced by the generator 
given a a span of sentence representation  $\mathcal{S}$. The first term of the objective trains the discriminator to assign high probability to a set $R_i\in \mathcal{Q} $ i.e  a set rules that were generated by Algorithm 1. The second term is trained such that the discriminator assigns low probability to a set  $\mathcal{P}$ from the generator. The smoothness $\mathcal{S}_p$ is added to encourage the generator to produce  similar rules for adjacent sentences. The gradient penalty $\mathcal{G}_p$  achieves stabilization of the discriminator by penalizing the gradient norm of the discriminator  with regards to the input   \cite{Gulrajani2017}.
\subsection{Decision Module.}
This module uses the trained GAN model to generate the most likely rule set $\mathcal{P}$ when a BERT representation of sentences in the conversation history is fed into the GAN model. The goal of this module is to establish the set $R_i\in \mathcal{Q}$ where the rules $\mathcal{P}$  generated by GAN are mapped to as shown in equation 6.
 \begin{equation}
  Sim(R_i,\mathcal{P})=|R_i\cap \mathcal{P}|
 \end{equation}
 where $i=1,2\cdots, size(\mathcal{Q})$.\\
$Sim(R_i,\mathcal{P})$ basically compares the one hot encodings  of  $\mathcal{P}$ and $R_i$ both having dimension $|\mathcal{U}|$. The set  $R_i\in \mathcal{Q}$ where  $ |R_i\cap \mathcal{P}|$  that has the highest overlap between  $R_i$ and $\mathcal{P}$ i.e $max(Sim(R_i,\mathcal{P}))$ is picked as the set  where the GAN generated set $\mathcal{P}$ belongs to.
To establish which rules have not been entailed by the conversation history, we perform a set difference 
\begin{equation}
setDiff(R_i,\mathcal{P})=R_i \setminus \mathcal{P}
\end{equation}
There are three key options that the module can adopt  based on the $Sim(R_i,\mathcal{P})$.  If  $\forall  R_i \in  \mathcal{Q}$, $|Sim(R_i,\mathcal{P})|=0$,  it means that the module has judged that the conversation history  does not match any rule set  $R_i\in \mathcal{Q}$ and it should generate "irrelevant" as the answer.  If the $max(Sim(R_i,\mathcal{P}))>0$  and $|setDiff(R_i,\mathcal{P})|=0$,  it means all rules of a set  $R_i\in \mathcal{Q}$ of a given subject are entailed by the conversation history and a definitive answer to the user inquiry should be generated by the system. Finally, if  $max(Sim(R_i,\mathcal{P}))>0$   and $|setDiff(R_i,\mathcal{P})|\geq1$,  it means some rules of a given subject are not entailed by the conversation history hence further inquiry by the system is necessary.
\section{ Answer Generation Module}
Here we fine tune  BERT to identify whether two sentences are a negation of each other. Therefore given two sentences BERT outputs 0 representing \textit{ not a negation} or 1 signaling that sentence $A$ is a \textit{negation} of $B$.
If rule set $ R_i$ is deemed to match $\mathcal{P}$ as described in section 4.1, we use the fine tuned BERT to detect if negation exists between sentences that generated the paired  rules. If negation exists in any pair of rules the system generates \textit{"No"} as an answer to the user. If no negation is detected in all paired rules, and $|setDiff(R_i,\mathcal{P})|\geq1$ the system invokes the question generation module to seek  further clarification on the rules that that are not entailed after which  the answer generation module is invoked again. If no negation is detected between paired rules and $|setDiff(R_i,\mathcal{P})|=0$   a \textit{"Yes"} answer is generated.

\section{Question Generation Module}
Given a rule $r_i$, this module seeks to create a natural question that seeks to clarify information related to the rule $r_i$. if  $|Sim(R_i,\mathcal{P})|>0$    and $|setDiff(R_i,\mathcal{P})|\geq1$ the system needs to clarify  some or all the rules in $setDiff(R_i,\mathcal{P})$. The number of rules in  in the set $|setDiff(R_i,\mathcal{P})|$  is the potential number of follow up question that the system will generate. Once a system has asked a question relating to a rule $r_i\in |setDiff(R_i,\mathcal{P})|$ the rule $r_i$ is removed from $|setDiff(R_i,\mathcal{P})|$ and the answer generation module is invoked taking into account the user's response to the inquiry.
\subsection{ Rule encoding}
For given  rule $r_i$ that the system needs to perform an inquiry on,   this module utilizes  the span of sentences $S_i$  that generated the rule $r_i$ according to  Algorithm 1. These  sentences,  are then  encoded by BERT as described  section 1. However, BERT is  now configured to return word embeddings as opposed to an embedding for the whole sentence.  After BERT encoding,  sentences $S_i$ that created the rule $r_i$  are now  represented as  tokens $x=\{x_1,x_2,\cdots,x_n\}$. The goal is to to generate a question $y=\{y_1,y_2\cdots, y_k\}$  given the tokens  $x=\{x_1,x_2,\cdots,x_n\}$. The task of this module can be framed as finding the most likely  question $\bar{y}$ such that:
\begin{equation}
\bar{y}=\underset{y}{\mathrm{argmin}} P(y|x)
\end{equation}
Here, $ P (y|x)$ is the conditional log-likelihood of
the predicted question sequence y, given the input
x. 
To generate the question using word level embeddings, we employ the technique proposed in \cite{Auli2016} \cite{Du2017} where the next word of  the question is predicted based on the input sentence and the current predicted word of the question as shown in equation 9.
\begin{equation}
P(y|x)=\prod_{i = 1}^{n}P(y_t|x,y_i)
\end{equation}
where $i<t$
\par Concretely,  we utilize  the Long Short-Term Memory (LSTM) network \cite{sema1997} to generate the question.
The hidden state of the recurrent network at time $t$
is computed based on the representation of  previous predicted word and previous hidden state $h_{t-1}$ as shown in equation 10. The initial hidden state $h_0$ is initiated as the representation of the  sentence $S_i$ generated by the BERT encoder.
\begin{equation}
h_t=LSTM(y_{t-1},h_{t-1})
\end{equation}
The prediction of a word $y_i$  belonging to the question is generated based on equation 11.
\begin{equation}
P(y_t|x,y<t)=\text{softmax}(tanh(W_s tanh(W_t[h_t;c_t]))
\end{equation}
where $W_s$ and $W_t$ are parameters to be learned during training while $c_t$ is the attention encoding of the input $x$ at time $t$.

\section{Experimental Setup}
\subsection{Dataset}
To evaluate our model we use ShARC dataset \cite{Saeidi2020}. We first construct an extensive rule text where we visit every unique  URL contained in the ShARC dataset extract all the relevant text contained in that web page then merge the text in the different web pages into a single continuous document. Text extracted from the different  websites are placed directly next to each. We discard most headings contained in the web pages. Bullet points in the web page are reconstructed into sentences. User scenario, user question and a concatenation  of system generated inquiry( follow up questions) and the related  user answer are placed directly after the relevant text as sentences. To extract  sentences from the  rule text we use spaCy\footnote{\url{https://spacy.io/}}.
\subsection{Sentence Encoding and Rule Extraction}
During the fine tuning of BERT to encode the sentences,  we use the following hyper-parameters:  a batch size of 32, a learning rate  of  $5e-5$. The BERT is fine-tuned with 100,000 steps and a warm-up of 10,000 steps

\subsection{Graph Construction And Partitioning}
To construct a graph $G$ ,we use the Gaussian similarity function in equation 12 to compute the similarity between sentences $S_i$ and $S_j$ representing the edges $v_i$ and $v_j$ respectively of the graph $G$.
\begin{equation}
Sim(S_i,S_{i+1})=\exp(-||S_i-S_j||^2)/(2\sigma^2)
\end{equation}
For graph partitioning in Algorithm 1, we varied the  value of $k$ based on the number of vertices $n$ on the graph $G$. We set $k=log(n)$. We found this as optimum value that provided a compromise between recall and precision when extracting rules in the rule text.
\subsection{Rule entailment setup}
For the GAN model, Both the Generator and Discriminator were trained  using Adam\cite{} optimizer with $\beta_1=0.5$ and $\beta_2=0.98$. The  weight decay  for the discriminator  was set to be $1e-4$. The generator and the discriminator  were trained with a learning rate of $1e-4$ and $1e-5$ respectively. The GAN model was trained  for a total of 100,000 steps. The optimizing during training is  alternated between discriminator and the generator hence both  the discriminator and generator is updated 50,000 times.
\par For the  generator,  we  set up a convolutional neural network (CNN) model proposed in \cite{Jacovi2019}  such that  the input of the convolutional layer is a BERT encoder  word level representation of a sentence $\mathcal{S}$  i.e for  the  words $\{w_1,w_2,\cdots,w_n\}\in \mathcal{S}$,  BERT encoder generates a matrix $\mathcal{M}\in R^{n\times d}$ where $d$ is the dimension of each  word $w_i$ generated by BERT.
The words $\{w_1,w_2,\cdots,w_n\}$ are fed into  the convolutional layer. A $k$ sized sliding window is then passed over the words. For every 
$u_i=\{w_i,\cdots,w_{i+k+1}\}\in R^{d\times k}
$ 
where
   $0\leq i\leq n-k$
   each $u_i$ is processed by a filter of similar dimension i.e $f_j \in R^{d\times k}$
  we use $m$ filters in the convolutional layer.
  The output $C\in R^{n\times M}$ of the convolutional layer  is a matrix of size $ n\times m$.  We then apply  max-pooling across the word dimension to generate a vector  $P\in R^{m}$  which is is then  fed into ReLU non-linearity. Finally, a linear fully connected layer $F\in R^{\mathcal{U}\times m}$ generates  the probability  distribution over the rules in the set $\mathcal{U}$ associated with the sentence $\mathcal{S}$,
 During implementation,  we experimented with  different filter   sizes and noted that $k=3$ produced the best results. We used  30 filters.

  For the discriminator, we used two  convolutional layers  followed by a single max  pooling layer followed by another  two  convolutional layers  followed by a max pooling layer. In all the convolutional layers we used 30 convolutional filters with a filter size of 5.  A dropout of 0.1 is used after the second max pooling function.  The discriminator takes as its  input a vector  of $|\mathcal{U}|$  dimension  which  represents  probability distribution over all the rules in the universal set $|\mathcal{U}|$ . The output is a single logit value which is an indication whether the sample is from the set $\mathcal{Q}$ or not.
  \section{Evaluation}
To evaluate the competitiveness  of the developed Unsupervised based Conversational Machine Reading Tool (UCMRT) , we perform a direct comparison  to several state of the art tools.\\
\subsection{Baseline}
\textbf{DISCERN} \cite{discourse2020} splits the rule text into  elemetary discourse units (EDU) using a pre-trained discourse segmentation model, it then trains a  supervised model  to predict whether each EDU is entailed by the user feedback in a conversation. Using the trained model the system gives a feedback to the user's question.\\
\textbf{$E^3$} \cite{E2020} proposes a number of threshold based heuristics that extracts rules from the rule text, checks for entailment of the extracted rules  and uses LSTM based model to generate follow up questions to clarify rules that have not been entailed.\\
\textbf{EMT} \cite{Gao2020} first encodes the conversational history using BERT, it then uses  explicit memory tracking that relies on recurrent network to update the entailment state of a rule sentence. For decision making, it exploits  entailment oriented reasoning based on the 
current states of rule sentences.\\
We also compare the results of the our developed CMR tool  to  \textbf{Seq2Seq} and \textbf{Pipeline} whose results are reported in \cite{Saeidi2020}

\subsection{Results}
The evaluation of the developed system is shown in table 1.
\begin{table}[ht]
\caption{Performance Comparison of UCMRT on blind held-out test of ShARC end to end task } 
\centering 
\begin{tabular}{c c c c c } 
\hline\hline 
\textbf{Model} & \textbf{BLEU1} & \textbf{BLEU 4} & \textbf{Micro Accuracy}&\textbf{Macro Accuracy}\\ [0.5ex] 
\hline 
$E^{3}$ & 54.1 & 38.7 & 73.3 & 67.6 \\ 
\textbf{Pipeline} & 54.4 & 34.4 & 61.9&68.9 \\
\textbf{Seq2Seq} & 34.0 & 7.8 & 44.8 &42.8 \\   \textbf{DISCERN}& 64.0 & 49.1 & 73.2&78.3 \\
\textbf{EMT}& 60.9 &46.0 	& 69.4 &74.8 \\ 
\textbf{UCMRT}& 66.7 &50.2 	& 76.5 &79.7  \\
\hline 
\end{tabular}
\label{table:n}
\end{table}
The results of \textbf{UCMRT} compared to other state of the art tools on held out test of ShARC is shown in table 1. \textbf{UCMRT} which is majorly based on unsupervised learning reports 3.3\% improvement on micro averaged accuracy  as compared to \textbf{DISCERN} which currently reports the highest value  of 73.2\%. Similarly, for macro averaged accuracy, \textbf{UCMRT} reports a 1.4\%  improvements as compared to \textbf{DISCERN} which currently reports the best macro-accuracy of 78.3\%.
Further, we  investigate the performance of \textbf{UCMRT} on each distinct class of answers given to the user. The results are shown in table 2
\begin{table}[ht]
\caption{Prediction accuracy of UCMRT on answer generation per class(Yes, No, Inquire and Irrelevant) on ShARC dataset } 
\centering 
\begin{tabular}{c c c c c } 
\hline\hline 
\textbf{Model} & \textbf{Yes} & \textbf{No} & \textbf{Inquiry} &\textbf{Irrelevant} \\
\hline 
$E^{3}$ & 65.9 & 70.6 & 60.5& 96.4\\ 
   \textbf{DISCERN}& 71.9 & 75.8 &73.3& 99.3\\
\textbf{EMT}& 70.5 &73.2 	& 70.8 &98.6\\ 
\textbf{UCMRT}& 74.1 &77.2 	& 76.5 &98.9\\
\hline 
\end{tabular}
\label{table:n}
\end{table}
 The results in table 1 and the significance improvement of the prediction accuracy of  \textbf{UCMRT} on the three classes i.e Yes,No and Inquiry demonstrates that \textbf{UCMRT} generator of the GAN  is able to extract majority of the rules in the rule text and the model  has better understanding of rule entailment and is able capture negation that exist between conversational history and rule text.
  \section{Ablation Study}
  Motivated by the evaluation done in  \cite{discourse2020},  where they compared the results when RoBERTa   encoder  is replaced with BERT encoder while the entire system remains the same, we also set up another version \textbf{UCMRT(RoBERTa)} where BERT is replaced with  RoBERTa in figure 3.
  The results is reported in table 3.
 \begin{table}[ht]
\caption{RoBERTa vs BERT } 
\centering 
\begin{tabular}{c c c  } 
\hline\hline 
\textbf{Model} & \textbf{Micro Acc} & \textbf{Macro Acc}  \\
\hline 
 \textbf{UCMRT(BERT)}& 76.5 & 79.7 \\ 
 \textbf{UCMRT(RoBERTa)}& 77.6 & 79.2\\
 
\hline 
\end{tabular}
\label{table:n}
\end{table}
 Based on the results presented in table 3, RoBERTa reports an improvement of 0.9\%  on micro-accuracy and 0.5\% degradation on the macro-accuracy. This shows that on the overall, replacing BERT encoder  with RoBERTa in the \textbf{UCMRT} tool has no significant impact on the performance of the tool.\\
\textbf{UCMRT} extracts a span of sentences that addresses a given subject(topic), then uses spectral partitioning technique to extract rules within a given subject. We investigated  if this technique presented a performance advantage as compared to simple sentence splitting i.e  that a a rule is composed within a given sentence 
 \begin{table}[ht]
\caption{Spectral rule extraction vs Sentence splitting } 
\centering 
\begin{tabular}{c c c  } 
\hline\hline 
\textbf{Model} & \textbf{Micro Acc} & \textbf{Macro Acc}  \\
\hline 
 \textbf{UCMRT}& 76.5 & 79.7 \\ 
 \textbf{UCMRT(Sentence Splitting)}& 73.6 & 74.2\\
  
\hline 
\end{tabular}
\label{table:n}
\end{table}
 From the results in table 4, performing a trivial sentence splitting significantly  degrades the performance of the tool. From our analysis we observed that when multiple rules are contained within a sentence, most rules are ignored and treated as a single rule hence when it comes to the entailment module  described in section 4 the generator of GAN fails to generate most of the rules which degrades the performance downstream.
 \section{Conclusion}
 This paper presents an unsupervised based CMR tool. The paper also looks into how CMR tool can be configured to answer more general questions from users. We specifically exploit spectral clustering algorithm to extract rules from the rule text and then we use the GAN unsupervised model to learn how the rules are extracted from rule text. The trained GAN model is able to generate rule(s) given a sentence from the rule text. We then apply   set theory to check for rule entailment. For question generation, we apply a simple LSTM model to generate a question to the user. The experiments based on the developed tool achieves state of the art results

\bibliographystyle{unsrtnat}
\bibliography{mybibfile}   

\end{document}